\documentclass[10pt,twocolumn,letterpaper]{article}

\usepackage{cvpr}
\usepackage{times}
\usepackage{epsfig}
\usepackage{graphicx}
\usepackage{amsmath}
\usepackage{amssymb}

\usepackage{multirow}
\usepackage{booktabs}
\usepackage{mathrsfs}
\usepackage{subfig}

\usepackage[pagebackref=true,breaklinks=true,letterpaper=true,colorlinks,bookmarks=false]{hyperref}

\cvprfinalcopy 

\def\cvprPaperID{0408} 

\ifcvprfinal\fi
\begin{document}

\title{Embedding Label Structures for Fine-Grained Feature Representation}

\author{Xiaofan Zhang\\
UNC Charlotte\\
Charlotte, NC 28223\\
{\tt\small xzhang35@uncc.edu}
\and
Feng Zhou\\
NEC Lab America\\
Cupertino, CA 95014\\
{\tt\small feng@nec-labs.com}
\and
Yuanqing Lin\\
NEC Lab America\\
Cupertino, CA 95014\\
{\tt\small ylin@nec-labs.com}
\and
Shaoting Zhang\\
UNC Charlotte\\
Charlotte, NC 28223\\
{\tt\small szhang16@uncc.edu}
}

\maketitle

\begin{abstract}
Recent algorithms in convolutional neural networks (CNN) considerably advance the fine-grained image classification, which aims to differentiate subtle differences among subordinate classes. However, previous studies have rarely focused on learning a fined-grained and structured feature representation that is able to locate similar images at different levels of relevance, \eg, discovering cars from the same make or the same model, both of which require high precision. In this paper, we propose two main contributions to tackle this problem. 1) A multi-task learning framework is designed to effectively learn fine-grained feature representations by jointly optimizing both classification and similarity constraints. 2) To model the multi-level relevance, label structures such as hierarchy or shared attributes are seamlessly embedded into the framework by generalizing the triplet loss.
Extensive and thorough experiments have been conducted on three fine-grained datasets, \ie, the Stanford car, the Car-333, and the food datasets, which contain either hierarchical labels or shared attributes. Our proposed method has achieved very competitive performance, \ie, among state-of-the-art classification accuracy when not using parts. More importantly, it significantly outperforms previous fine-grained feature representations for image retrieval at different levels of relevance. 
\end{abstract}

\section{Introduction}

\begin{figure}[t]
\begin{center}
   \includegraphics[width=0.99\linewidth]{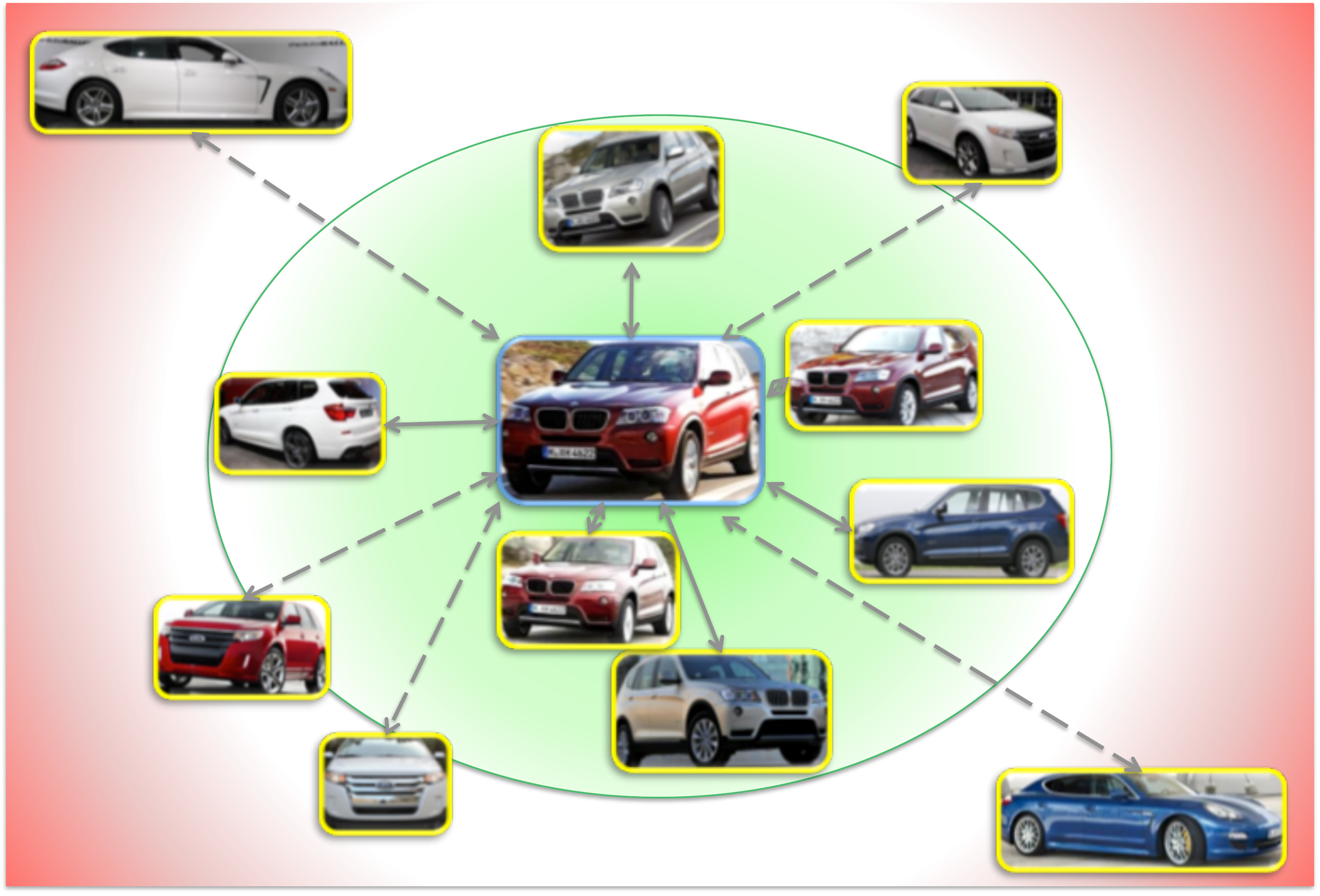}
\end{center}
   \caption{Examples from a fine-grained car dataset~\cite{krause2013collecting}, where the similarity can be defined at different levels, \ie, body type, model, and even viewpoint, indicated by the distance to the query in the center. Images within the circle have exactly the same fine-grained labels, \ie, make and model, and the closest two also share the same viewpoint.}
\label{fig:motivation}
\end{figure}

Recent advances in image understanding (\eg, classification, detection, segmentation, retrieval) have been driven by the success of convolutional neural networks (CNN)~\cite{jia2014caffe,krizhevsky2012imagenet,szegedy2014going,simonyan2014very,sermanet2013overfeat}. Particularly, models of fine-grained image categorization have made tremendous progress in recognizing subtle differences among subordinate classes, such as different models of cars~\cite{krause2013collecting,krause20133d,lin2014jointly,yang2015large}, breeds of animals~\cite{khosla2011novel,parkhi2012cats,deng2013fine,berg2014birdsnap,krause2015fine,lin2015deep,lin2015bilinear}, and types of food dishes~\cite{bossard2014food,yang2010food}. Most of previous methods focus on improving the classification accuracy, by learning critical parts that can align the objects and discriminate between neighboring classes~\cite{yang2012unsupervised,chai2013symbiotic,berg2013poof,zhang2013deformable,zhang2014part,goering2014nonparametric}, or using distance metric learning to alleviate the issue of large intra-class variation~\cite{weinberger2009distance,wah2014similarity,wang2014learning,qian2015fine}. However, such studies have rarely been dedicated to learn a structured feature representation that can discover similar images at different levels of relevance. Fig.~\ref{fig:motivation} shows examples of similar cars from a fine-grained dataset~\cite{krause2013collecting}. Having the same fine-grained labels indicate exactly the same make, model and year, while cars are still similar even they have different labels, \eg, the same make but different year, or the same body style (\eg, SUV, Coupe) from different make. Such hierarchy of similarity should also be explored in fine-grained feature representation, since it is applicable to various use cases such as the recommendation of relevant products in e-commerce. 

To obtain the fine-grained feature representation, a potential solution is to incorporate similarity constraints (\eg, contrastive information~\cite{chopra2005learning} or triplets~\cite{norouzi2012hamming,chechik2010large}). For example, Wang et al.~\cite{wang2014learning} proposes a deep ranking model to directly learn the similarity metric by sampling triplets from images. However, these strategies still have several limitations in fine-grained datasets: 1) Although the features learned from triplet constraints are effective at discovering similar instances, its classification accuracy may be inferior to the fine-tuned deep models that emphasize on the classification loss, as demonstrated in our experiments. In addition, the convergence speed using such constraints is usually slow. 2) More importantly, previous methods for fine-grained features do not embed label structures, which is critical to locate images with relevance at different levels.

In this paper, we propose two contributions to solve these issues: 1) A multi-task deep learning framework is designed to effectively learn the fine-grained feature representation without sacrificing the classification accuracy. Specifically, we \emph{jointly optimize} the classification loss (\ie, softmax) and the similarity loss (\ie, triplet) in CNN, which can generate both categorization results and discriminative feature representations. 
2) Furthermore, based on this framework, we propose to seamlessly \emph{embed label structures} such as hierarchy (\eg, make, model and year of cars) or attributes (\eg, ingredients of food).
We evaluate our methods on three fine-grained datasets, \ie, the Stanford car, the Car-333, and a fine-grained food dataset, containing either hierarchical labels or shared attributes.
The experimental results demonstrate that our feature representation can precisely differentiate fine-grained or subordinate classes, and also effectively discover similar images at different levels of relevance, both of which are challenging problems.

The rest of the paper is organized as follows. Section \ref{sec:rw} provides a brief review of fine-grained image categorization and the recent approaches of learning fine-grained feature representation. Section \ref{sec:m} introduces our method which learns feature representation by multi-task learning and embedding label structures. Experiments are presented in Section \ref{sec:e}, and we conclude the paper in Section \ref{sec:c}.

\section{Related Work}
\label{sec:rw}

Fine-grained image understanding aims to differentiate subordinate classes.
Its main challenges are the following: 1) Many fine-grained classes are highly correlated and are difficult to distinguish due to their subtle differences, \ie, small inter-class variance. 2) On the other hand, the intra-class variance can be large, partially due to different poses and viewpoints. Many methods have been proposed to alleviate these two problems. In this section, we emphasize on the methods that are most relevant to our approaches, particularly the ones on fine-grained feature representation.

Many algorithms have been proposed to leverage parts of objects to improve the classification accuracy. Part based models~\cite{yang2012unsupervised,chai2013symbiotic,berg2013poof,zhang2013deformable,zhang2014part,goering2014nonparametric,xiao2014application} are proposed to capture the subtle appearance differences in specific object parts and reduce the variance caused by different poses or viewpoints. Different from these part-based methods, distance metric learning can also addresses these challenges by learning an embedding such that data points from the same class are clustered together, while those from different classes are pushed apart from each other. In addition, it ensures the flexibility of grouping the same category, such that only a portion of the neighbors from the same class need to be pulled together. For example, Qian et al.~\cite{qian2015fine} proposed a multi-stage metric learning framework that can be applied in large-scale high-dimensional data with high efficiency.
In addition to directly classify the images using CNN, it is also possible to generate discriminative features that can be used for classification. In this context, DeCAF~\cite{icml2014c1_donahue14} is a commonly used feature representation with promising performance achieved by training a deep convolutional architecture on an auxiliary large labeled object database. These features are from the last few fully connected layers of CNN, which have sufficient generalization capacity to perform semantic discrimination tasks using classifiers, reliably outperforming traditional hand-engineered features.

One limitation of the above mentioned methods is that they are essentially driven by the fine-grained class labels for classification, while it is desired to incorporate similarity constraints as well. Therefore, other than using classification constraints alone (\eg, softmax),  several similarity constraints have been proposed for feature representation learning.
For example, siamese network~\cite{chopra2005learning} defines similar and dissimilar image pairs, with the requirement that the distance between dissimilar pairs should be larger than a certain margin, while the one from similar pairs should be smaller.  This type of similarity constraint can effectively learn feature representations for various tasks, especially for the verification~\cite{wolf2014deepface,sharma2015scalable}.
An intuitive improvement is to combine the classification and the similarity constraints together for better performance. This is particularly relevant to our framework. For example, \cite{sun2014deep,yi2014learning,bell2015learning} proposed to combine the softmax and contrastive loss in CNN via joint optimization. It improved traditional CNN 
because contrastive constraints might augment the information for training the network. Different from these approaches, our method leverages the triplet constraint~\cite{norouzi2012hamming,chechik2010large} instead of the contrastive ones, since triplet can preserve the intra-class variation~\cite{schroff2015facenet}, which is critical to the learning of fine-grained feature representation. Note that triplet constraint has been used in feature learning~\cite{wang2014learning,lai2015simultaneous,wah2014similarity}, face representation~\cite{schroff2015facenet}, and person re-identification~\cite{ding2015deep}. Particularly, there are also efforts on combining this with the softmax. A representative example is that \cite{parkhideep} proposed to learn a face classify first, and then use the triplet constraint to fine-tune and boost the performance. It achieved promising accuracy in face recognition. Although we also integrate triplet information with the traditional classification objective, our method jointly optimizes these two objectives simultaneously, which is different from \cite{parkhideep}. As shown in the experiments, this joint optimization strategy generates better feature representations for fine-grained image understanding. In addition, our framework can also easily support the embedding of label structures in a unified framework, \eg, hierarchy or shared attributes, which have been proven useful in various studies ~\cite{berg2010automatic,duan2012discovering,akata2013label,vedaldi2014understanding,xie2014hyper,yu2014fine,chen2015deep}, but not well explored in learning fine-grained feature representation that can model similarity at different levels.


\section{Methodology}
\label{sec:m}

\begin{figure*}[ht]
\begin{center}
\includegraphics[width=0.99\linewidth]{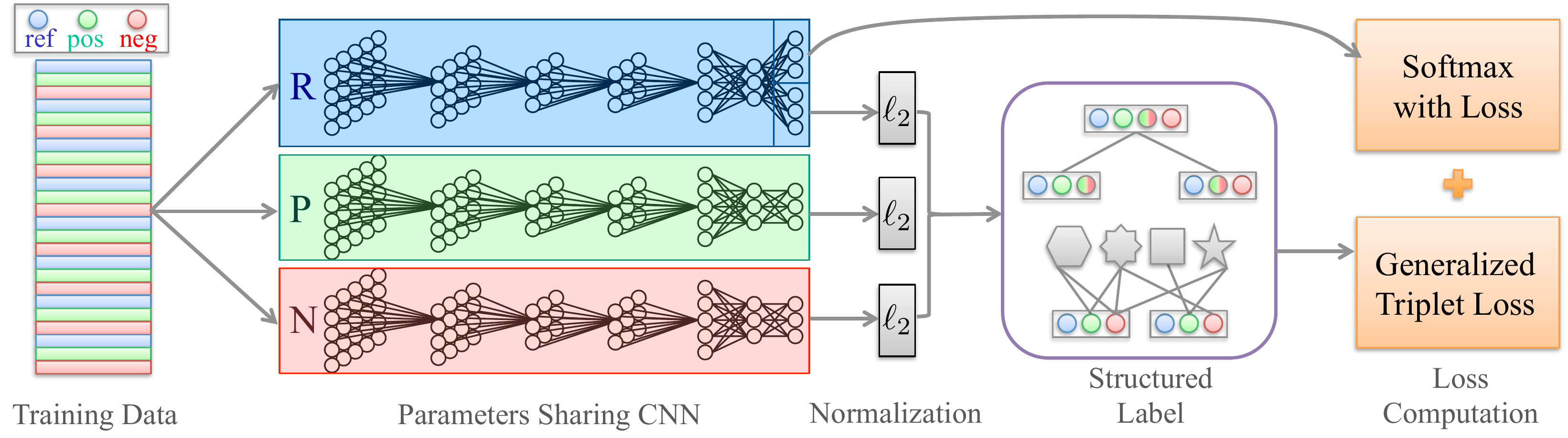}
\end{center}
\caption{Our framework takes the triplets (\ie, the reference, the positive and the negative images) and the label of the reference image as the input, which pass through the three networks with shared parameters. The label structures are embedded in the loss layer, including the hierarchy or shared attributes. Two types of losses are optimized jointly to obtain the fine-grained classifier and also the feature representation.}
\label{fig:framework}
\end{figure*}

\subsection{Jointly Optimize Classification and Similarity Constraints}

Traditional classification constraints such as softmax with loss are usually employed in CNN for fine-grained image categorization, which can distinguish different subordinate classes with high accuracy. Suppose that we are given $N$ training images
$\{r_i, l_i\}_{i=1}^N$ of $C$ classes,
where each image $r_i$ is labeled as class $l_i$. Given the output of the last fully connected layer $f_s(r_i, c)$ for each class $c=1,\cdots, C$, the loss of softmax can be defined as the sum of the negative log-likelihood over all training images $\{r_i\}_i$:
\begin{align}
E_s(r,l) = \frac{1}{N} \sum\limits_{i=1}^N - \log \underbrace{\frac{e^{f_s(r_i, l_i)}}{\sum_{c=1}^C e^{f_s(r_i, c)}}}_{P(l_i | r_i)}, \label{eq:softmax}
\end{align}
where $P(l_i | r_i)$ encodes the posterior probability of the image $r_i$ being classified as the $l_i$th class. In a nutshell, Eq.~\ref{eq:softmax} aims to ``squeeze" the data from the class into a corner of the feature space. Therefore, the intra-class variance is not preserved, while such variance is essential to discover both visually and semantically similar instances.

To address these limitations, we explicitly model the similarity constraint in CNN using a multi-task learning strategy. Specifically, the triplet loss is fused with the classification objective as the similarity constraint. A triplet consists of three images, denoted as $(r_i, p_i, n_i)$, where $r_i$ is the reference image from a specific class, $p_i$ an image from the same class, and $n_i$ an image from a different class.
Given an input image $r_i$ (similarly for $p_i$ and $n_i$), this triplet-driven network can generate a feature vector $f_t(r_i) \in \mathbb{R}^{D}$, where the hyper-parameter $D$ is the feature dimension after embedding. Ideally, for each reference $r_i$, we expect its distance from any $n_i$ of different class is larger than $p_i$ within the same class by a certain margin $m>0$, \ie,
\begin{align}
\mathscr{D}(r_i, p_i) + m < \mathscr{D}(r_i,n_i), \label{eq:triplet}
\end{align}
where $\mathscr{D}(\cdot, \cdot)$ is the squared Euclidean distance between two $\ell_2$-normalized vectors $f_t(\cdot)$ of the triplet network. To enforce this constraint in CNN training, a common relaxation~\cite{norouzi2012hamming} of Eq.~\ref{eq:triplet} can be defined as the following hinge loss:
\begin{align}
\begin{split}
&E_t(r, p, n, m) = \\
&\frac{1}{2N} \sum_{i=1}^N \max \lbrace 0,\mathscr{D}(r_i,p_i) - \mathscr{D}(r_i, n_i)+ m \rbrace. \label{eq:triplet_loss}
\end{split}
\end{align}

In the feature space defined by $f_t(\cdot)$, it can group the $r$ and $p$ together while repelling the $n$ by minimizing $E_t(r,p,n,m)$.
The gradient can be computed as:
\small
\vspace{-.5em}
\begin{align}
\triangledown W_t &= 2(f_t(r_i) - f_t(p_i)) \frac{\partial f_t(r_i) - \partial f_t(p_i)}{\partial W_t} \nonumber \\
& \quad -2(f_t(r_i) - f_t(n_i)) \frac{\partial f_t(r_i) - \partial f_t(n_i)}{\partial W_t},
\\[-1.5em]\nonumber
\end{align}
\normalsize
if $\mathscr{D}(r_i, n_i) -\mathscr{D}(r_i, p_i) < m$, otherwise $0$.
Different from the pairwise contrastive loss~\cite{chopra2005learning} that forces the data of the same class to stay close with a fixed margins, the triplet loss allows certain degrees of intra-class variance. Despite its merits in learning feature representation, minimizing Eq.~\ref{eq:triplet_loss} for recognition tasks still  has several disadvantages. For example, given a dataset with $N$ image, the number of all possible triplets is $N^3$,
and each triplet contains much less information (\ie, similar or dissimilar constraints with margins) compared with the classification constraint that provides a specific label among $C$ classes. This can lead to slow convergence. Furthermore, without the explicit constraints for classification, the accuracy of differentiating classes can be inferior to the traditional CNN using softmax, especially in fine-grained problems where the differences of subordinate classes are very subtle.

Given the limitations of training with the triplet loss (Eq.~\ref{eq:triplet_loss}) solely, we propose to jointly optimize two types of losses using a multi-task learning strategy. Fig.~\ref{fig:framework} shows the CNN architecture of our joint learning. The $R,P,N$ networks share the same parameters during training. After the $\ell_2$ normalization, the outputs of the three networks (\ie, $f_t(r), f_t(p), f_t(n)$) are transmitted to the triplet loss layer to compute the similarity loss $E_t(r, p, n, m)$. In the meantime, the output of the network $R$, $f_s(r)$, is forwarded to the softmax loss layer to compute the classification error $E_s(r,l)$. Then, we integrate these two types of losses through a weighted combination:
\begin{align}
E = \lambda_s E_s(r,l) + (1-\lambda_s) E_t(r, p, n, m), \label{eq:combined_loss}
\end{align}
where $\lambda_s$ is the weight to control the trade-off between two types of losses. We optimize Eq.~\ref{eq:combined_loss} using the standard stochastic gradient descent with momentum. 
The final gradient is computed as a $\lambda$-weighted combination of $\triangledown W_s$ from the classification constraint and $\triangledown W_t$ from the similarity constraint, and propagated back to lower layers.
This framework of unifying three networks through Eq.~\ref{eq:combined_loss} not only learns the discriminative features but also preserves the intra-class variance, without sacrificing the classification accuracy. In addition, it resolves the issue of the slow convergence when only using the triplet loss.  Regarding the sampling strategy, one can either follow the methods in Facenet~\cite{schroff2015facenet}, or employ hard mining approaches to explore challenging examples in the training data. Both of them are effective in our framework, since jointly optimizing $E_s(r,l)$ facilitates the searching of good solutions, allowing certain flexibility for the sampling.  

During the testing stage, this framework takes one image as an input, and generates the classification result through the softmax layer, or the fine-grained feature representation after the $\ell_2$ normalization. This discriminative feature representation can be employed for various tasks such as classification, verification and retrieval, which is more effective than solely optimizing the softmax with loss.

\subsection{Embed Label Structures}

As discussed before, an effective feature representation should be able to search relevant instances at different levels (\eg, Fig.~\ref{fig:motivation}), even not within the same fine-grained class. Our framework serves as a baseline
to naturally embed label structures, without sacrificing the classification accuracy on fine-grained datasets. In particular, we aim to handle two types of label structures, \ie, hierarchical labels and shared attributes, both of which have wide applications in practice.

\subsubsection{Generalized Triplets for Hierarchical Labels}

\begin{figure}[h]
\begin{center}
\includegraphics[width=0.99\linewidth]{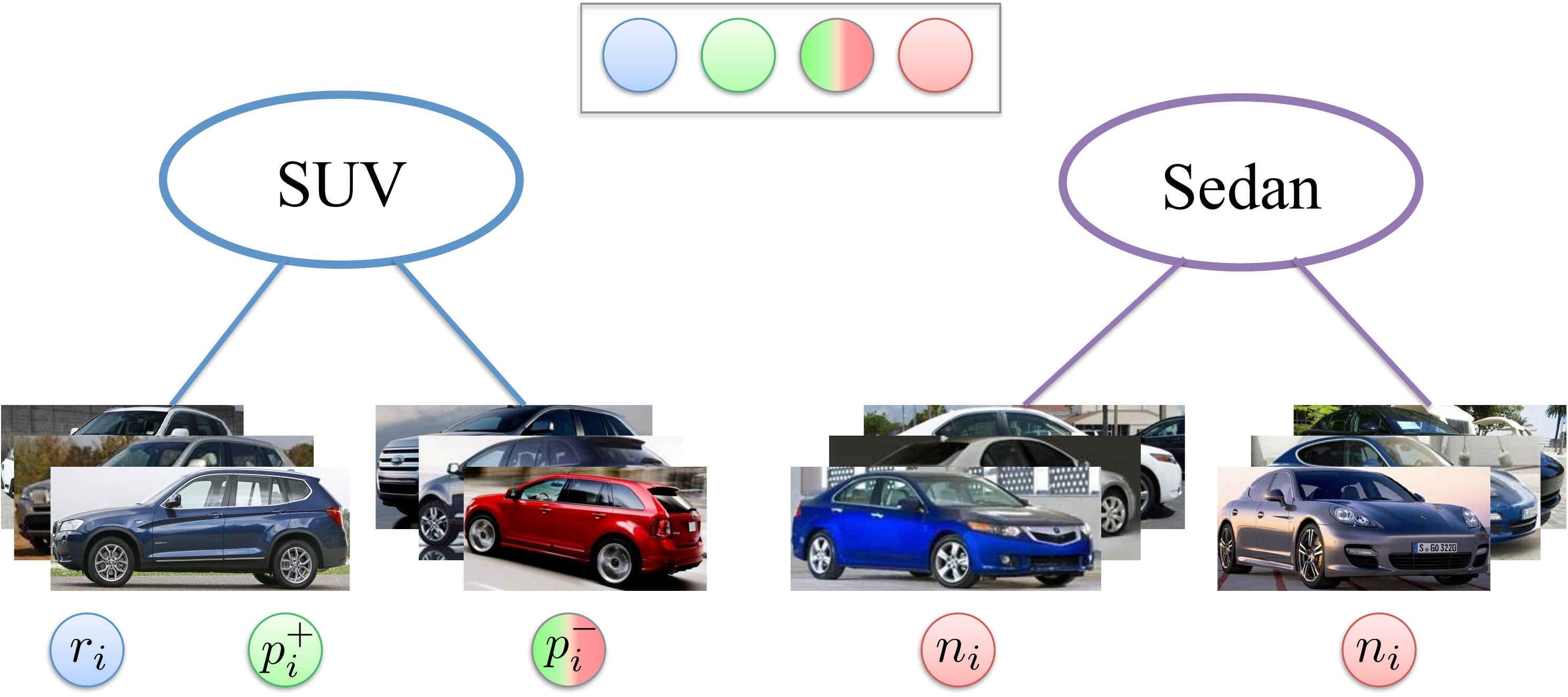}
\end{center}
\caption{The hierarchy of labels in the fine-grained car dataset~\cite{krause2013collecting}. Blue ($r_i$) means the reference image, green ($p^+_i$) denotes the image with the same fine-grained label (\ie, the same make, model and year), green-red ($p^-_i$) represents different fine-grained labels but the same coarse label (\ie, the body type), and red ($n_i$) indicates different coarse labels.}
\label{fig:hier}
\end{figure}

In the first case, the fine-grained labels can be naturally grouped in a tree-like hierarchy based on semantics or domain knowledge. The hierarchy can contain multiple levels. 
For simplicity purpose, we explain the algorithm with a two-level structure, and then generalize to multiple levels. Fig.~\ref{fig:hier} illustrates an example of two-level labels from a car dataset~\cite{krause2013collecting}, 
where the fine-grained car models in the leaf nodes are grouped according to their body types in the roots.

To model this hierarchy of coarse and fine class labels, we propose to generalize the concept of triplet. Specifically, \emph{quadruplet} is introduced to model the two-level structure. Each quadruplet, $(r_i, p^+_i, p^-_i, n_i)$, consists of four images. Similar to triplet, $p^+_i$ denotes the image of the same fine-grained class as the reference $r_i$. The main difference is that in quadruplet, all negative samples are classified into two sub-categories: the more similar one $p^-_i$ that shares the same coarse class with $r_i$, and the more different one $n_i$ sampled from different coarse classes. Given a quadruplet, this hierarchical relation among the four images can be described in two inequalities,
\begin{align}
\mathscr{D}(r_i, p^+_i) + m_1 < \mathscr{D}(r_i, p^-_i) + m_2 < \mathscr{D}(r_i,n_i), \label{eq:quad}
\end{align}
where the two hyper-parameters, $m_1$ and $m_2$, satisfying $m_1 > m_2 > 0$, control the distance margins across the two levels. It is worth to mention that if Eq.~\ref{eq:quad} is satisfied, then $\mathscr{D}(r_i, p^+_i) + m_1 + m_2 < \mathscr{D}(r_i, n_i)$ automatically holds. Compared to triplet, quadruplet is able to model much richer label structures between different levels, \ie, coarse labels and fine-grained labels. As a result, the learned feature representation can discover relevant instances that are appropriate in specific scenarios, \eg, locating a car with specific model and year, or finding SUVs from different body types.

Regarding the sampling strategy, all training images are used as the references in every epoch. For each reference image $r_i$, we select $p^+_i$, $p^-_i$ and $n_i$ from other corresponding classes, depending on both fine and coarse labels.
To incorporate this quadruplet constraint in CNN training, we propose to decompose Eq.~\ref{eq:quad} into two triplets, $(r_i, p^+_i, p^-_i)$ and $(r_i, p^-_i, n_i)$, phrased as \emph{generalized triplets}.
Similar to Eq.~\ref{eq:triplet_loss}, our approach seeks for the optimal parameters that minimize the joint loss over the sampled quadruplets:
\begin{align}
&E_q(r, p^+, p^-, n, m_1, m_2) = \nonumber\\
& \frac{1}{2N} \sum_{i=1}^N \max\lbrace 0,\mathscr{D}(r_i,p^+_i) - \mathscr{D}(r_i,p^-_i) + m_1-m_2\rbrace \nonumber\\
& \quad + \frac{1}{2N} \sum_{i=1}^N \max\lbrace 0, \mathscr{D}(r_i,p^-_i) - \mathscr{D}(r_i,n_i) + m_2\rbrace.
\end{align}
Clearly, this generalized triplets can be naturally incorporated into our multi-task learning framework (Eq.~\ref{eq:combined_loss}).

So far we have mainly discussed in the scenario of a two-level label hierarchy, through the generalized triplet representation of quadruplet. In fact, our method is also applicable to the more general multi-level case using the same strategy, \ie, representing a ``tuplet" with generalized triplets.
Similar to the quadruplet sampling strategy, each tuplet is formed by selecting the classes at different similarity levels, from which training images are sampled (one image at each level).
Therefore, a tuplet from an $x$-level hierarchy contains $x+2$ images (e.g., the quadruplet from a two-level hierarchy has four images).
This tuplet is decomposed into $x$ triplets, by taking the reference image and two more images from two adjacent levels.
Intuitively, this means that multiple triplets are sampled to represent different levels of similarity, \ie, images with the same finer-level labels are more similar than ones sharing the same coarser-level labels.
Same as the two-level case, it can be optimized using the multi-task learning framework based on triplets.
Even though this is not exhaustive sampling or exact decomposition for the tuplet,  the generalized triplets are representative enough to ensure a good performance, which is demonstrated in our experiments (Section~\ref{sec:car333}).
It is also worth mentioning that the traditional triplet is a special case of the generalized triplet, \ie, only one-level hierarchy.

\subsubsection{Generalized Triplets for Shared Attributes}

\begin{figure}[t]
\begin{center}
\includegraphics[width=0.99\linewidth]{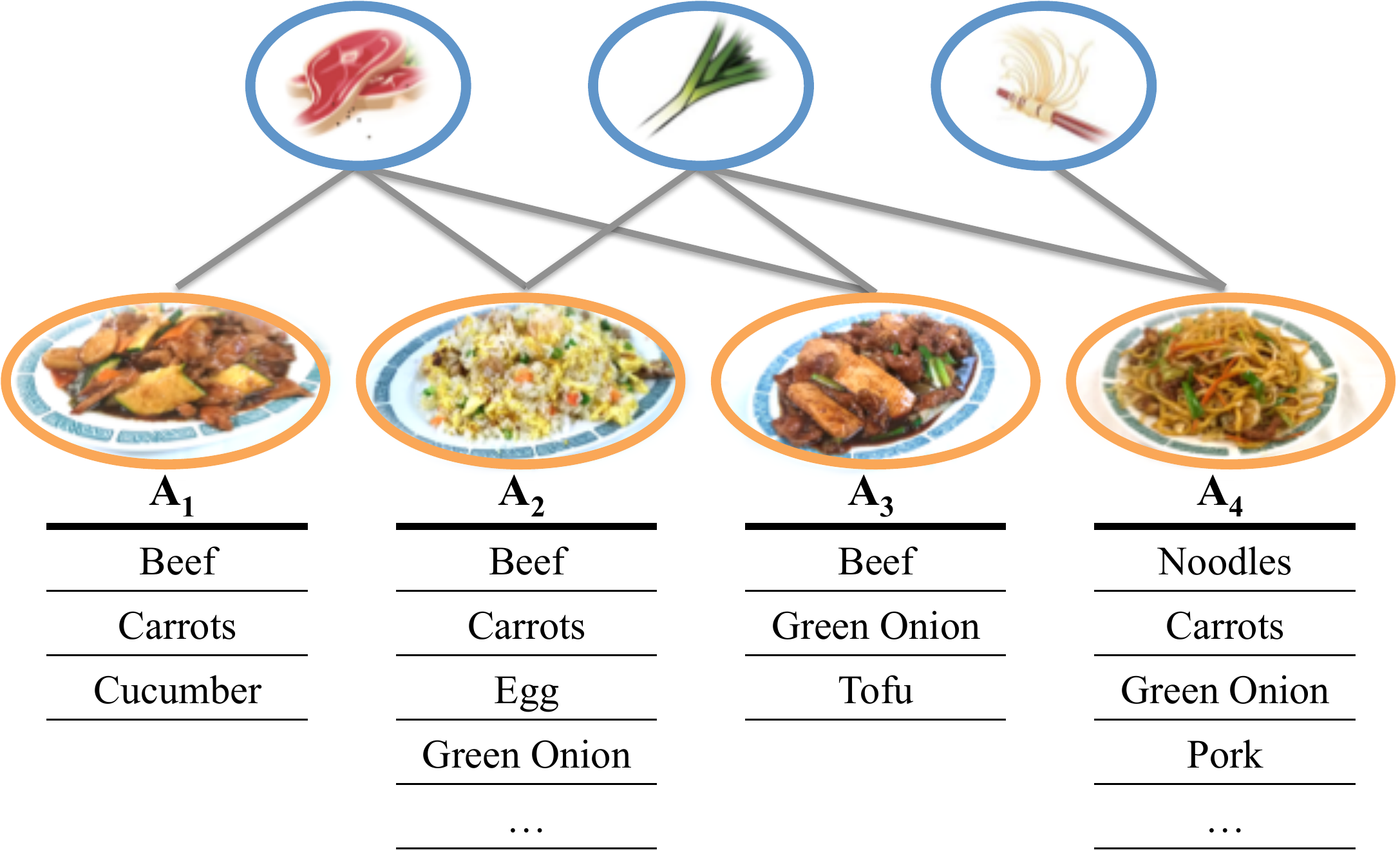}
\end{center}
   \caption{The shared attributes in our food dataset, where the attributes ($A_1$-$A_4$) mean the ingredients. }
\label{fig:attri}
\end{figure}

In the second case, fine-grained objects can share common attributes with each other. For instance, Fig.~\ref{fig:attri} illustrates that fine-grained food dishes can share some ingredients,
indicating relevance at different levels. Intuitively, classes that share more attributes should be more similar than the classes sharing less attributes.
Unlike the tree-like hierarchy in the first case, we are not able to directly model the label dependency as Eq.~\ref{eq:quad}, because some fine-grained classes can own multiple attribute labels. Instead, we model this graph dependency using a modified triplet idea. To have a better understanding of our method, we can consider the first three dishes shown in Fig.~\ref{fig:attri}. Although both the second and third dishes belong to different classes compared to the first one, the second dish shares more attributes (beef, carrots) with the first dish. This difference in attribute overlapping inspires us to re-define the margin $m$, \ie, the distance between $\mathscr{D}(r_i,p_i)$ and $\mathscr{D}(r_i,n_i)$, as the Jaccard similarity~\cite{Jaccard:Phytologist1912} of attributes from different classes:
\begin{align}
m = m_b \left(1 - \frac{|A_p \cap A_n|}{|A_p \cup A_n|}\right), \label{eq:jaccard}
\end{align}
where $m_b$ is a constant factor specified as the base margin, $A_p$ and $A_n$ are the sets of attributes belonging to the positive and negative categories, respectively. Therefore, the more attributes these classes share, the smaller margin this triplet has. Using such adaptive margin for the triplet loss, the learned feature can discover images containing common attributes as the query images. Similarly, Eq.~\ref{eq:jaccard} can be naturally incorporated in our multi-task learning framework based on the triplet loss. In fact, the original triplet constraint is also a special case of the multi-attribute constraint, when each fine-grained label only connects to one attribute.


\section{Experiments}
\label{sec:e}

\begin{figure}[t]
\centering
\includegraphics[width=1\linewidth]{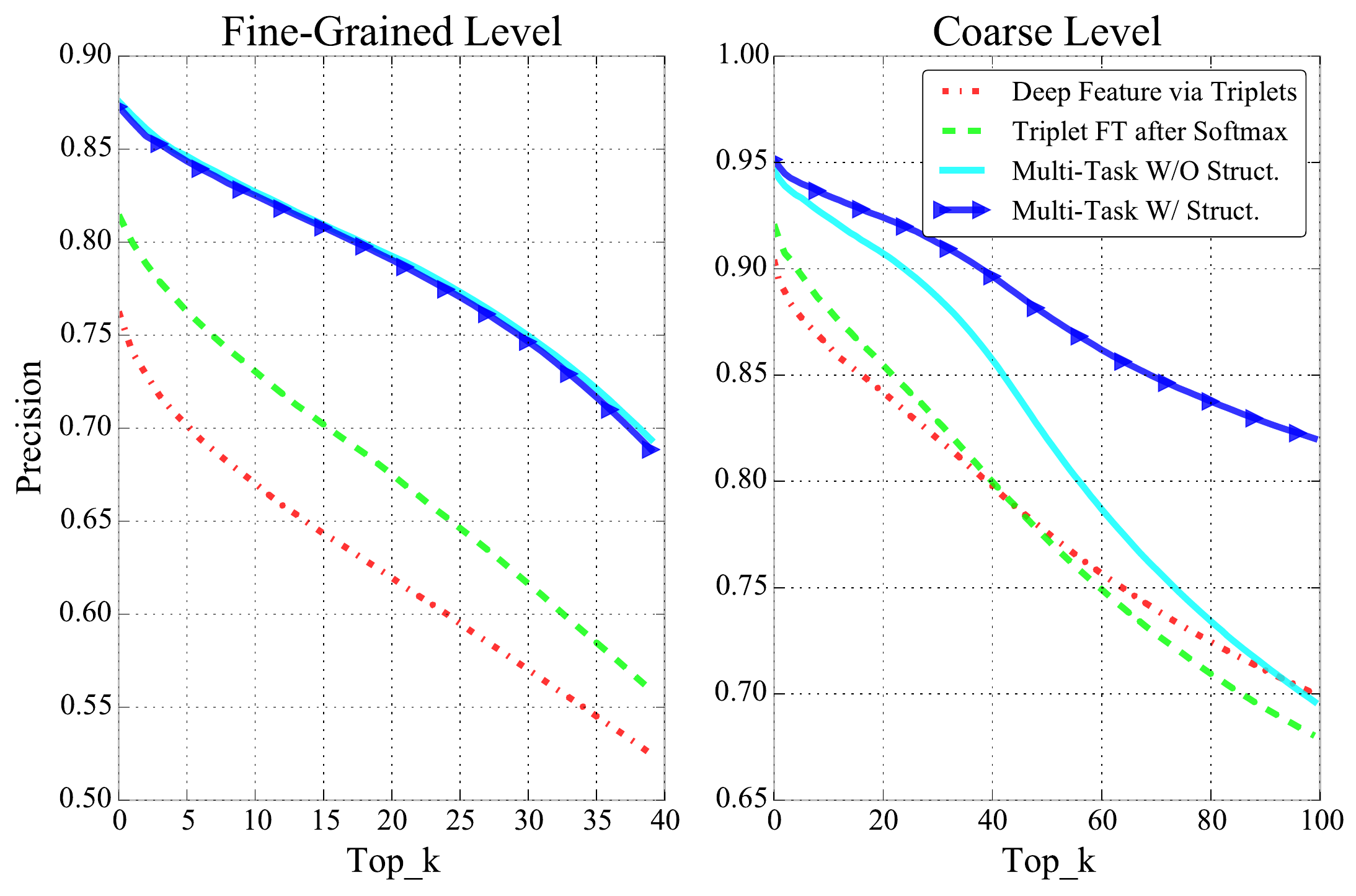}
\caption{Comparison of retrieval precision on the Stanford car, with two levels of labels. }
\label{fig:prec_carS}
\end{figure}

In this section, we conduct thorough experiments to evaluate this proposed framework on three fine-grained datasets with label structures,
Particularly, we aim to demonstrate that our learned feature representations can be used to retrieve images at different levels of relevance, with significantly higher precision than other CNN-based methods. In addition, we also report its promising classification accuracy on these fine-grained classes.

We focus on the comparison of four methods that can generate fine-grained feature representation:
1) deep feature learning by triplet loss~\cite{schroff2015facenet,wang2014learning}, 2) triplet-based fine-tuning after softmax~\cite{parkhideep}, \ie, not joint optimization, 3) our multi-task learning framework, and 4) our framework with label structures. In the classification task, besides these four methods, we also report the accuracy of using CNN with traditional softmax. All CNNs are based on GoogleNet~\cite{szegedy2014going}, and are fine-tuned on these fine-grained datasets for the best performance and fair comparisons. We also carefully follow the specifications from these compared papers for their settings and parameters.  Regarding our hyper-parameters, we empirically set the feature dimension as $200$, the margin as $0.2$, and the weight $\lambda_s$ as $0.8$, with discussions of the sensitivity in Section~\ref{sec:discussions}.

\subsection{Stanford Car with Two-Level Hierarchy }

The first experiment focuses on the efficacy of embedding hierarchical labels, using the Stanford car dataset~\cite{krause2013collecting}.
It contains 16,185 images (with bounding boxes) of 196 car categories, with 8,144 for training and the rest for testing.
The categories, \ie, fine-grained class labels, are defined as make, model and year, such as Audi S4 Sedan 2012.
Following~\cite{krause2013collecting}, we have assigned each fine-grained label to one of nine coarse body types, such as SUV, Coupe and Sedan (Fig.~\ref{fig:hier} in~\cite{krause2013collecting}), resulting in a two-level hierarchy.

\begin{figure}[t]
\centering
\subfloat[Without Label Structures]{\includegraphics[width=0.5\linewidth]{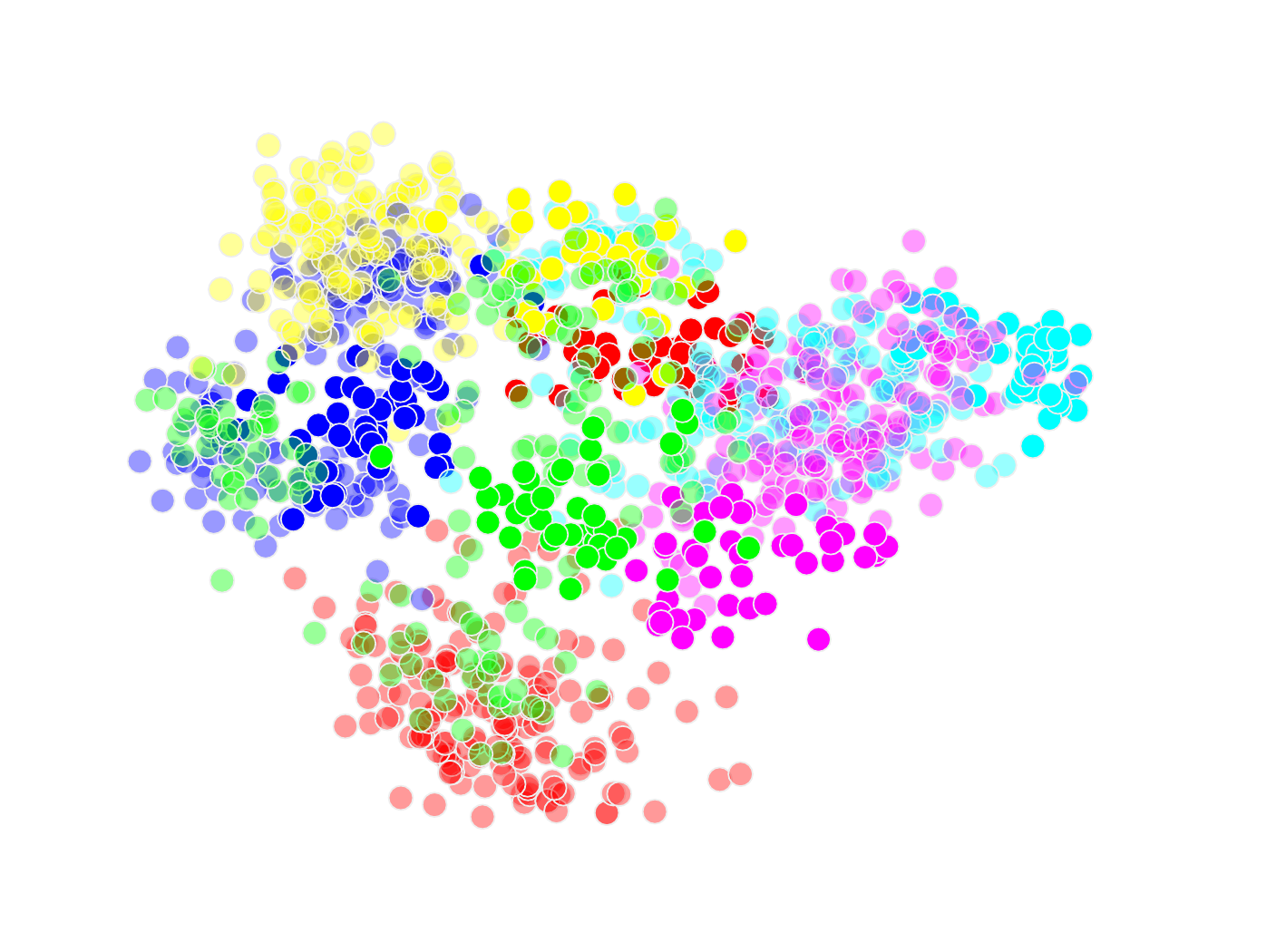}\label{fig:feat_tri}}
\subfloat[With Label Structures]{\includegraphics[width=0.5\linewidth]{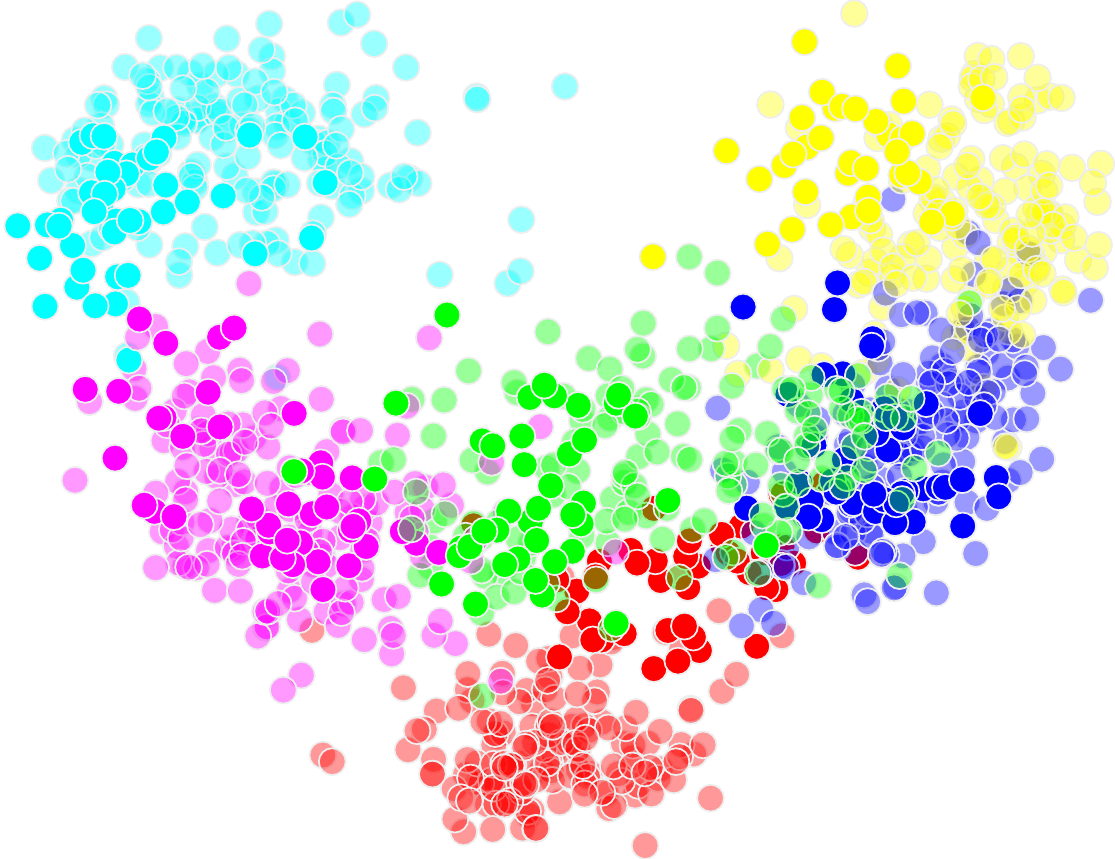}\label{fig:feat_quad}}
\caption{Visualization of features after dimension reduction. Different colors represents different coarse-level labels, and intensities (or transparency) from the same color indicate fine-grained labels. }
\label{fig:feat}
\end{figure}

Fig.~\ref{fig:prec_carS} shows the retrieval precision using feature representations extracted by various CNNs, at both the fine-grained level and the coarse level. At the fine-grained level, results from our multi-task learning methods are better than the others, \ie, at least $13.5\%$ higher precision at top-$40$ retrievals (using top-$40$ since each fine-category has around $40$ images). The reason is that the joint optimization strategy leverages the similarity constraints via triplets, which can augment the training information, assisting the network to reach better solutions.
No matter using the traditional or generalized triplets (\ie, without or with the label structures) in our framework, the difference of precision is within $0.5\%$, which can be caused by the sampling strategies.
At the coarse level, our method without label structures also fails to achieve high precision at top-$100$ retrievals, while using generalized triplets significantly outperforms the others, \ie, at least $12.4\%$ higher precision, demonstrating the efficacy of our embedding scheme.
To provide insights of our promising results on this coarse-level retrieval, we extract features from our multi-task learning framework using traditional and generalized triplets, and visualize them in Fig.~\ref{fig:feat} after dimension reduction. Six coarse-level classes are randomly chosen, and five fine-level classes are sampled from each coarse one. The features from generalized triplets are consistently much better separated than ones from traditional triplets, benefited from the embedding of label structures.

We also report the classification accuracy of these methods on fine-grained classes. A fine-tuned GoogleNet achieves $86.9\%$. Learning deep features via triplets alone~\cite{schroff2015facenet,wang2014learning} attains $78.7\%$, which is worse than GoogleNet. The reason is that softmax with loss can explicitly minimize the classification error, while triplets attempt to implicitly separate classes by constraining the similarity measures.
Fine-tuning with triplets after the softmax~\cite{parkhideep} also aims to integrate the classification and similarity constraints, same as ours. This identification and verification framework achieves promising performance in face recognition. However, different from our framework, it embeds the triplet loss after learning a face classifier, \ie, not a joint optimization strategy as ours. This may adversely affect the classification accuracy in fine-grained image categorization, since triplet loss only implicitly constrains the classification error, which may not be sufficient in further differentiating subordinate classes during fine-tuning. As a result, it achieves $83.0\%$, which is worse than the fine-tuned GoogleNet.
Our multi-task learning framework achieves $88.4\%$ when jointly optimizing both types of losses,
which are higher than these compared methods, and among state-of-the-art results that do not use parts.

\subsection{Car-333 with Three-Level Hierarchy}
\label{sec:car333}

\begin{figure}[t]
\centering
\includegraphics[width=1\linewidth]{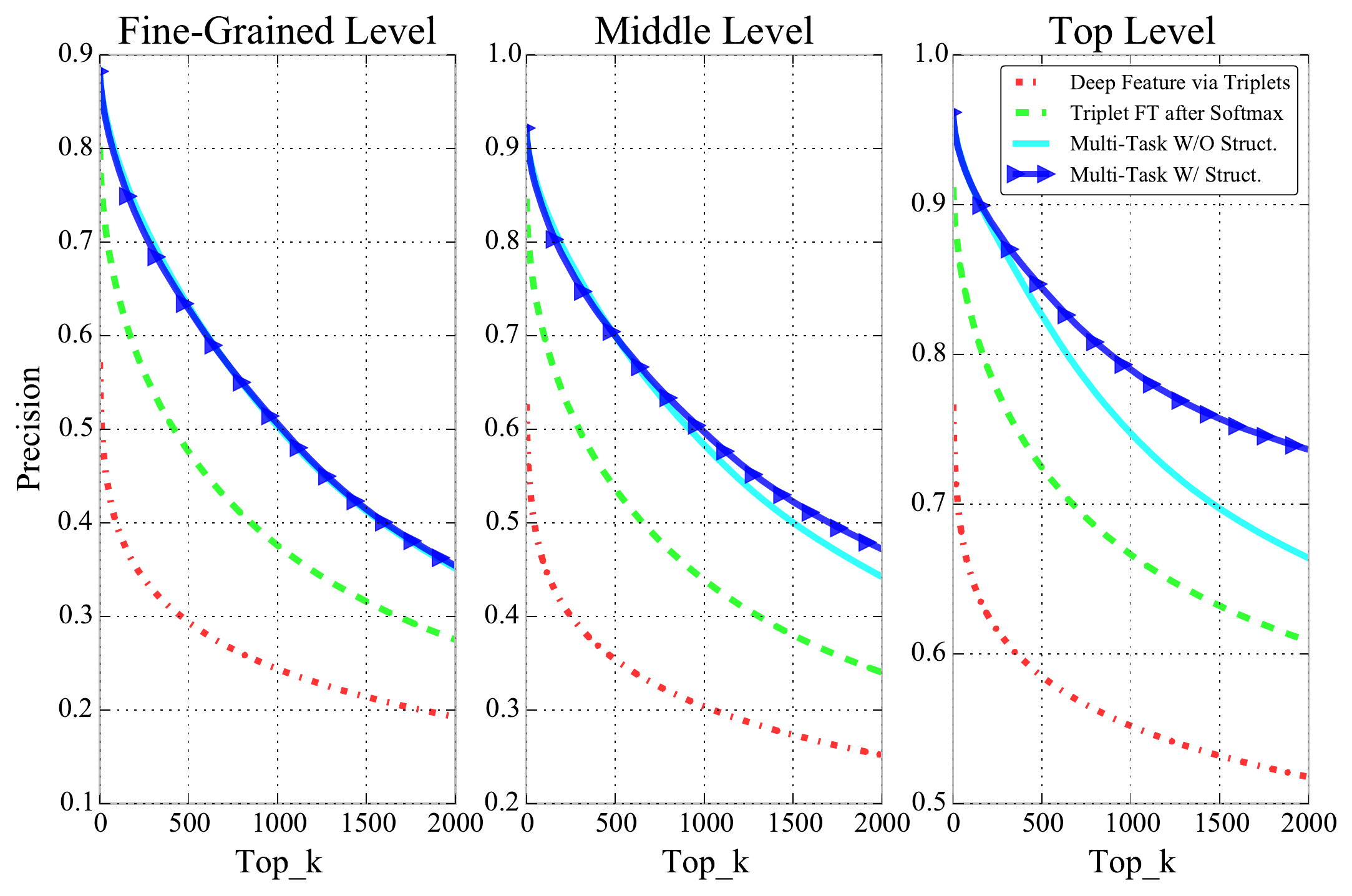}
\caption{Comparison of retrieval precision on the Car-333 dataset. Top-level means the car make only. Mid-level represents both make and model. Fine-level denotes the fine-grained labels of make, model and year range. }
\label{fig:prec_car}
\end{figure}

The second experiment also investigates the hierarchical labels, but using a much larger car dataset~\cite{xie2014hyper} to validate the scalability. These are end-user photos from the Craigslist, so they are more naturally photographed.
It contains 157,023 training images and 7,840 testing images, from 333 car categories. The categories are defined by make, model and year range. Note that two cars of the same model but from different years are considered as different classes.
The bounding boxes are generated by Regionlets~\cite{wang2013regionlets}, which produces promising results in car detection.
Different from the Stanford car, this has a three-level hierarchy: 333 fine-grained labels are grouped into 140 models by ignoring the difference of years, and then five makes (\ie, Chevrolet, Ford, Honda, Nissan, Toyota).

Fig.~\ref{fig:prec_car} shows the retrieval precision at these three levels. Since the training data is around 20 times larger than the previous one, we show the precision upon top-$2000$ retrievals (note that the number of images in a fine-level class can be less than $2000$). The results are consistent with the ones on the Stanford car, demonstrating that the strategy of generalized triplets is applicable to multi-level hierarchies. Specifically, our method with label structures is at least $13.2\%$ better than other methods in terms of the top-$2000$ retrieval precision at the middle level, and $12.8\%$ better at the top level. This is also $7.2\%$ better than ours without embedding structures at the top level, proving the efficacy of our generalized triplets.
In addition, such promising results also demonstrate that the scalability of our methods such as generalized triplets is sound.
Regarding the classification accuracy, GoogleNet achieves $87.9\%$, the deep feature via traditional triplets attains $61.2\%$, fine-tuning with triplets after softmax reaches $81.7\%$. It is worth mentioning that the deep feature via triplets has considerably worse performance on this dataset, compared to the results on the Stanford car. It indicates that this method does not have good scalability for fine-grained image categorization, although it is proven to be effective for other tasks such as verification and ranking~\cite{schroff2015facenet,wang2014learning}. On the other hand, jointly optimizing the softmax with loss can alleviate this issue even on this larger-scale dataset, as it directly tackles the classification problem. Using this strategy, our method achieves $89.4\%$, which is among state-of-the-art.

\subsection{Food Dataset with Shared Attributes}

\begin{figure}[t]
\centering
\includegraphics[width=1\linewidth]{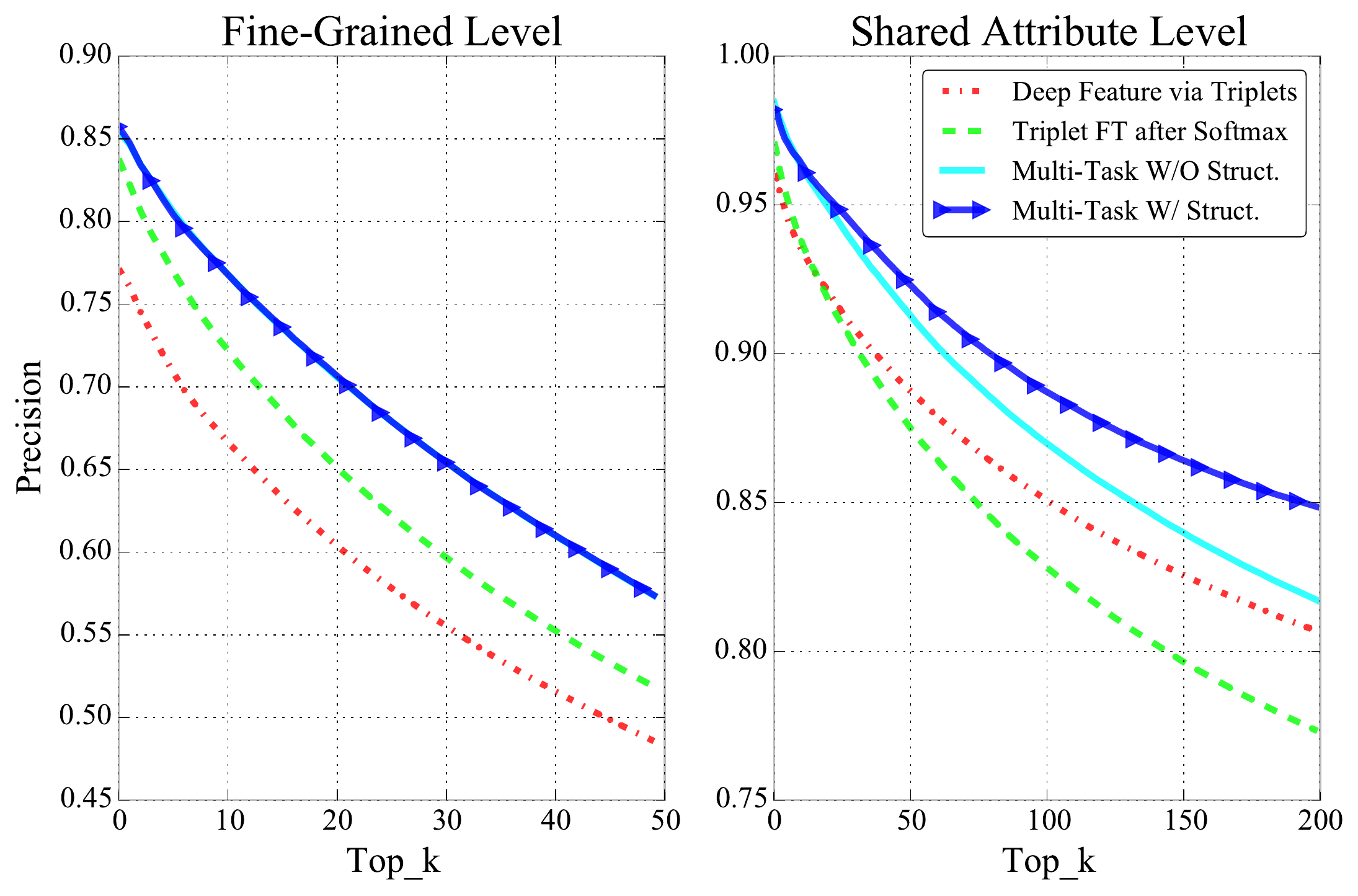}
\caption{Comparison of retrieval precision on the food dataset. Share Attribute Level means that two images are relevant if they share at least one attribute.}
\label{fig:prec_food}
\end{figure}

The third experiment aims to examine the embedding of shared attributes, using our newly collected food dataset that consists of ultra-fine-grained classes and rich class relationships. To generate this dataset, we sent multiple data collectors to six restaurants, and they took photos of most dishes during two months. In total, we acquired 37,086 food photos from 975 menu items, \ie, fine-grained class labels. In addition, we built a list of 51 ingredients, \ie, shared attributes, to precisely describe these dishes. This dataset is divided into 32,135 training and 4,951 testing images, and testing images are collected on different days from the training, to mimic a realistic scenario by avoiding potential correlations of taking photos in the same day (\eg, multiple photos from the same dish at the same time cannot be used for both training and testing).

Fig.~\ref{fig:prec_food} shows the retrieval precision on this food dataset with respect to top-$50$ retrievals, as each category has around $20$ to $50$ images. In addition to evaluate on the fine-grained labels, we also define a new level of relevance: two images are similar when they share at least one attribute.
Our method by embedding shared attributes outperforms the others by $5.5\%$ at the fine-grained level, and $4.2\%$ at the attribute level in terms of the precision. Since the precisions of these methods are already above $80\%$, such improvement means a reducing of $21.7\%$ for the errors.
Compared to our method without embedding attributes, it is nearly the same performance at the fine-grained level, while $3.1\%$ better at the attribute level (reducing errors by $16.9\%$), demonstrating the efficacy of the generalized triplets with adaptive margins.
Note that the improvement may not be as significant as on the other two datasets using hierarchical labels. The reason is that the similarity measure for attributes is more subtle, \ie, two cars having different coarse labels could be more distinguishable than two dishes sharing no attributes.
In terms of the classification accuracy, we have achieved $89.0\%$, comparing to $87.1\%$ by GoogleNet, $78.2\%$ by learning the deep feature and $86.1\%$ by fine-tuning with triplets after softmax. This is also a promising result, considering that this challenging dataset is ultra-fine-grained.

\subsection{Discussions}
\label{sec:discussions}

\begin{figure}[t]
\centering
\includegraphics[width=0.99\linewidth]{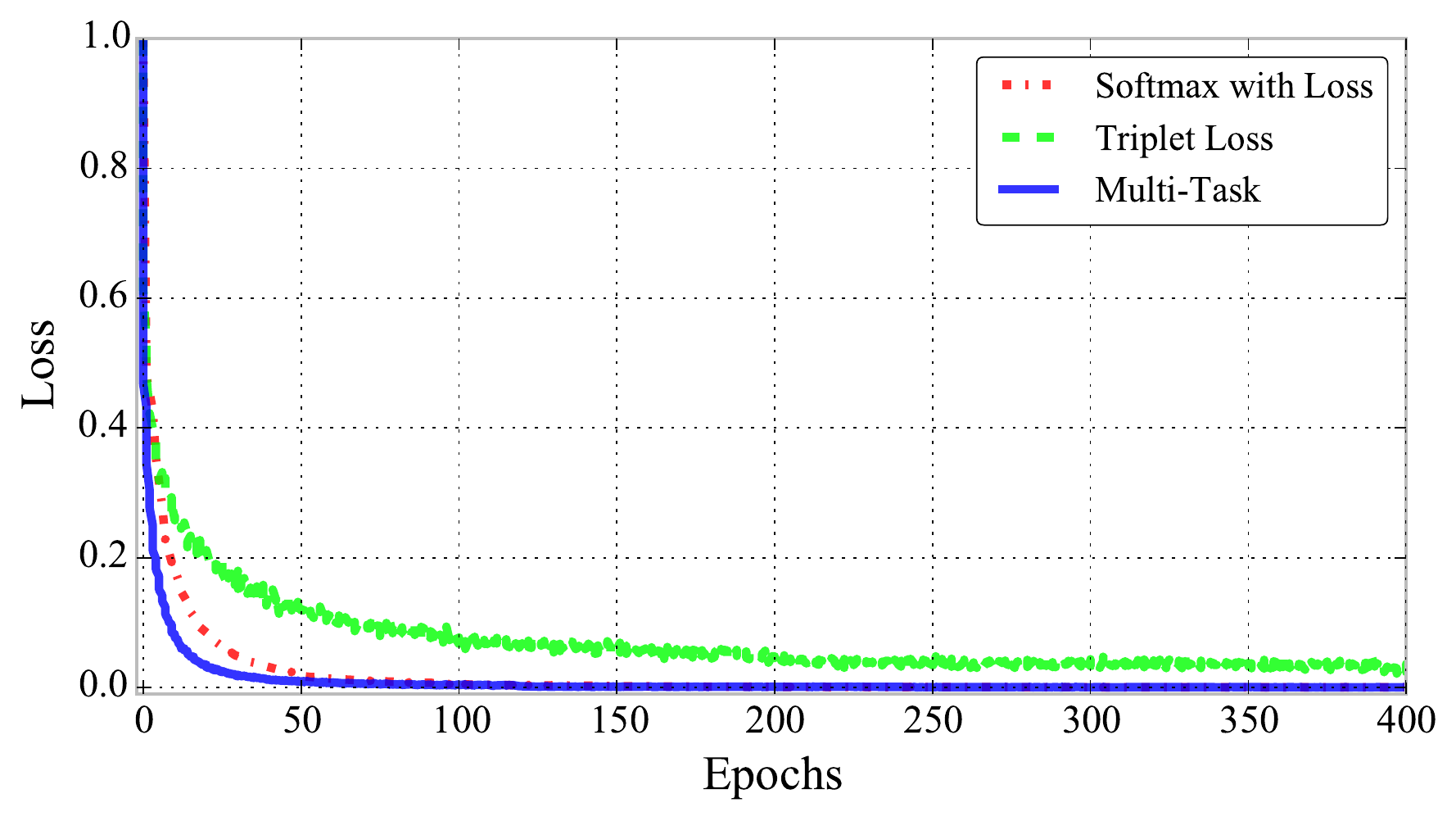}
\caption{Comparison of the convergence rate on the Stanford car dataset. The first $400$ epoches are shown for better visualization.}
\label{fig:converge}
\end{figure}

Fig.~\ref{fig:converge} shows the convergence rate of these methods. Since each triplet contains much less information compared to the one of using the label directly (\ie, softmax with loss), their convergence rates can be dramatically different. Particularly, using softmax with loss has much faster convergence rate than using triplet loss. Our multi-task learning framework jointly minimizes both of them, so it harvests augmented information from both sides, resulting in a fast convergence rate as well. Overall, our methods converge after $800$ epochs on the Stanford car, $150$ epochs on the Car-333, and $600$ epochs on the food dataset, which are reasonably fast in practice.

Our framework has one important parameter, the weight $\lambda_s$ to balance two types of losses, and setting $\lambda_s$ to be $0$ or $1$ degenerates our framework to deep feature learning by triplet loss~\cite{schroff2015facenet,wang2014learning} or GoogleNet (softmax with loss), respectively, which will either fail to differentiate fine-grained classes or lose the ability to generate effective feature representations. Since softmax with loss may contain more information than a triplet in each iteration, it is reasonable to assign a higher weight to softmax, \ie, larger than $0.5$. Our experiments show that the performance is not sensitive to small variations to $\lambda_s$, \ie, within $0.8\%$ difference in a range of $[0.55,0.85]$. Besides the weight, the feature dimension and the margin is also relevant to the classification accuracy.
From our extensive experiments, we observe that our methods are also stable with respect to their variations up to a certain range, \eg, within $2\%$ difference for feature dimensions from $128$ to $512$. Therefore, it is relatively easy to tune the hyper-parameters in our framework. In fact, we use the same group of parameters on all datasets.

\section{Conclusion}
\label{sec:c}

In this paper, we proposed a multi-task learning framework to effectively generate fine-grained feature representations by embedding label structures, such as hierarchical labels or shared attributes. In our method, the label structures are seamlessly embedded in CNN through the proposed generalized triplets, which can incorporate the similarity constraints at different levels of relevance. Such a framework retains the classification accuracy for subordinate classes with subtle differences, and at the same time considerably improves the image retrieval precision at different levels of label structures on three fine-grained datasets, including a newly-collected benchmark dataset for food. These merits warrant further investigating the embedding of label structures for learning fine-grained feature representation.

\pagebreak
{\small
\bibliographystyle{ieee}
\bibliography{abrv,egbib}
}

\end{document}